\documentclass[times,twocolumn,final]{elsarticle}

\usepackage{ctable}
\usepackage{color}
\usepackage{longtable}

\usepackage{caption}
\usepackage{subcaption}
\usepackage{medima_arxiv}
\usepackage{framed,multirow}
\usepackage{graphicx}
\usepackage{multirow}
\usepackage{amsmath}
\usepackage{multicol}
\usepackage{url}
\usepackage{xcolor}
\usepackage{moresize}
\usepackage{booktabs}
\usepackage{colortbl}
\usepackage{arydshln}
\usepackage[normalem]{ulem}
\usepackage[colorlinks=true,citecolor=blue,urlcolor=blue]{hyperref}

\usepackage{pifont}
\journal{ArXiv}
\definecolor{changes}{RGB}{0, 0, 0}

\begin{document}

\verso{Joël L. Lavanchy \textit{et~al.} Multi-centric phase and step recognition in LRYGB surgeries}  
\begin{frontmatter}

\title{\textcolor{changes}{Challenges in} Multi-centric Generalization: Phase and Step Recognition in Roux-en-Y Gastric Bypass Surgery}

\author[1,2,3]{Joël L. \snm{Lavanchy}\fnref{fn1}\corref{cor1}}
\author[3,4,5]{Sanat \snm{Ramesh}\fnref{fn1}}
\fntext[fn1]{Joël L. Lavanchy and Sanat Ramesh contributed equally and share co-first authorship.}
\author[5]{Diego \snm{Dall'Alba}}
\author[3,6]{Cristians \snm{Gonzalez}}
\author[5]{Paolo \snm{Fiorini}}
\author[1,2]{Beat \snm{Müller-Stich}}
\author[7]{Philipp C. \snm{Nett}}
\author[8]{Jacques \snm{Marescaux}}
\author[3,6]{Didier \snm{Mutter}}
\author[3,4]{Nicolas \snm{Padoy}}
\cortext[cor1]{Corresponding author: 
\textit{email}: joel.lavanchy@clarunis.ch (Joël L. Lavanchy)
}

\address[1]{University Digestive Health Care Center - Clarunis, Basel 4002, Switzerland}
\address[2]{Department of Biomedical Engineering, University of Basel, Allschwil 4123, Switzerland}
\address[3]{Institute of Image-Guided Surgery, IHU Strasbourg, Strasbourg 67000, France}
\address[4]{ICube, University of Strasbourg, CNRS,Strasbourg 67000, France}
\address[5]{Altair Robotics Lab, University of Verona, Verona 37134, Italy}
\address[6]{University Hospital of Strasbourg, Strasbourg 67000, France}
\address[7]{Department of Visceral Surgery and Medicine, Inselspital Bern University Hospital, Bern 3010, Switzerland}
\address[8]{IRCAD France,Strasbourg 67000, France}

\received{1 May 2013}
\finalform{10 May 2013}
\accepted{13 May 2013}
\availableonline{15 May 2013}
\communicated{S. Sarkar}

\begin{abstract}
\textbf{Purpose:} Most studies on surgical activity recognition utilizing Artificial intelligence (AI) have focused mainly on recognizing one type of activity from small and mono-centric surgical video datasets. It remains speculative whether those models would generalize to other centers. 
\\
\textbf{Methods:} In this work, we introduce a large multi-centric multi-activity dataset consisting of 140 surgical videos (MultiBypass140) of laparoscopic Roux-en-Y gastric bypass (LRYGB) surgeries performed at two medical centers, i.e., the University Hospital of Strasbourg, France (StrasBypass70) and Inselspital, Bern University Hospital, Switzerland (BernBypass70). The dataset has been fully annotated with phases and steps by two board-certified surgeons. Furthermore, we assess the generalizability and benchmark different deep learning models for the task of phase and step recognition in 7 experimental studies: 
1) Training and evaluation on BernBypass70\textcolor{changes}{;} 
2) Training and evaluation on StrasBypass70\textcolor{changes}{;} 
3) Training and evaluation on the joint MultiBypass140 dataset\textcolor{changes}{;}
4) Training on BernBypass70, evaluation on StrasBypass70\textcolor{changes}{;} 
5) Training on StrasBypass70, evaluation on BernBypass70\textcolor{changes}{;} 
6) Training on MultiBypass140, evaluation on BernBypass70\textcolor{changes}{;} 
7) Training on MultiBypass140, evaluation on StrasBypass70. 
\\
\textbf{Results:} The model's performance is markedly influenced by the training data. The worst results were obtained in experiments 4) and 5) confirming the limited generalization capabilities of models trained on mono-centric data. The use of multi-centric training data, experiments 6) and 7), improves the generalization capabilities of the models, bringing them \textcolor{changes}{beyond} the level of \textcolor{changes}{independent} mono-centric training and validation (experiments 1) and 2)). 
\\
\textbf{Conclusion:} MultiBypass140 shows considerable variation in surgical technique and workflow of LRYGB procedures between centers. Therefore, generalization experiments demonstrate a remarkable difference in model performance. These results highlight the importance of multi-centric datasets for AI model generalization to account for variance in surgical technique and workflows. The dataset and code are publicly available at \hyperlink{https://github.com/CAMMA-public/MultiBypass140}{https://github.com/CAMMA-public/MultiBypass140}.

\end{abstract}

\end{frontmatter}
\thispagestyle{empty}

\section{Introduction}

\begin{figure*}[t]
     \centerline{
     \includegraphics[width=\linewidth, trim={0cm {.07\linewidth} 0cm {.05\linewidth}}, clip]{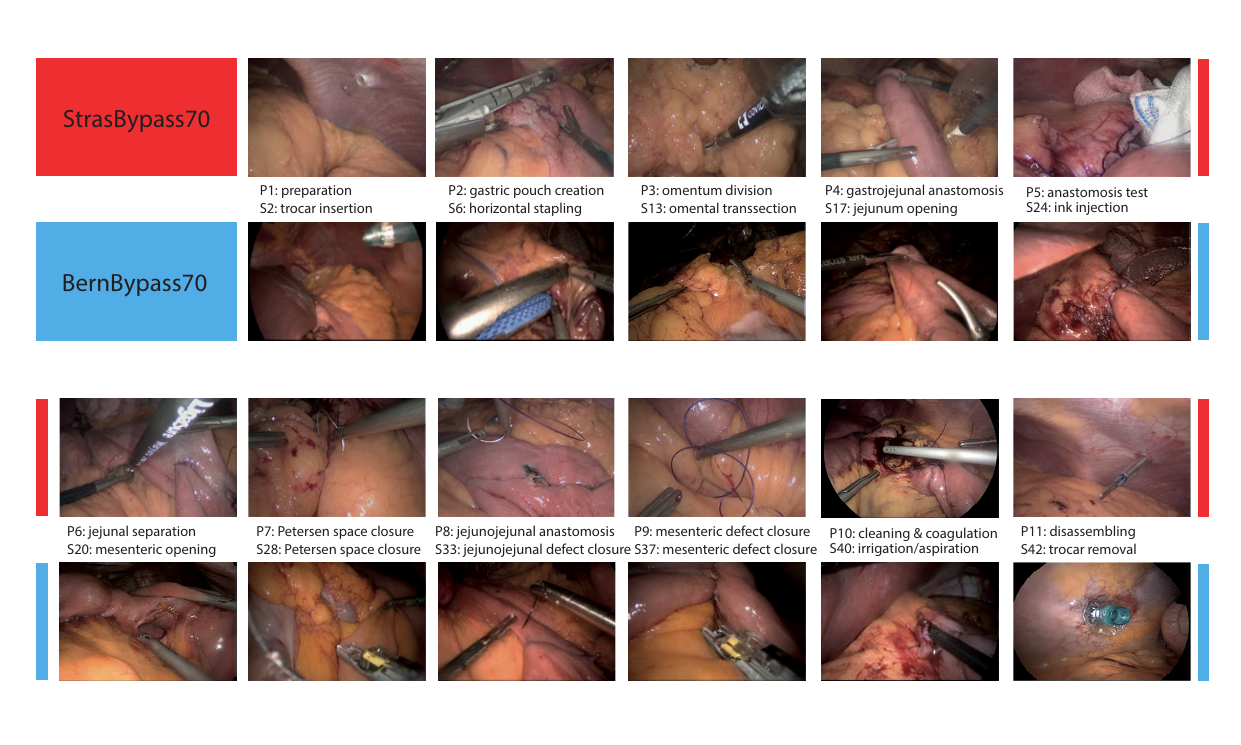}}
     \includegraphics[angle=90, width=0.49\linewidth, clip]{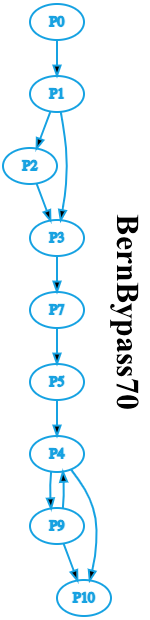}
     \includegraphics[angle=90, width=0.49\linewidth, clip]{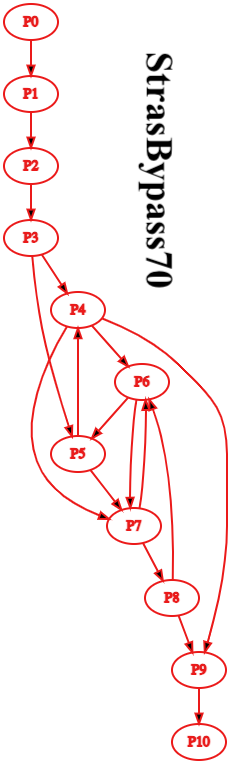}
     \caption{MultiBypass140 (Top): Sample video frames from Strasbourg University Hospital (StrasBypass70) and Inselspital Bern University Hospital (BernBypass70). (Bottom) Surgical workflow (modeled as phases~\citep{lavanchy2023proposal}) followed in more than 10 surgeries in each medical center.}
     \label{fig:sample_images}
\end{figure*}

The emerging field of Surgical Data Science (SDS) aims to impact the quality of interventional healthcare by collecting, organizing, analyzing, and modeling surgical data~\citep{MaierHein2022}. A principal element of SDS is analyzing intraoperative data collected in the operating room (OR) to model surgical workflows which eventually could improve patient outcomes by providing intraoperative assistance, streamlining surgical training~\citep{Lavanchy2021, Pedrett2023}, preoperative planning, and postoperative analysis. Additionally, knowledge about surgical workflows could be vital in developing autonomous surgical robotic platforms~\citep{Battaglia2021}. 

Research in SDS has proposed systematic decomposition of workflows\textcolor{changes}{'} multi-level activities - such as the whole procedure, phases, stages, steps, and actions~\citep{Katic2015, SAGESMeireles2021} - and developed various methods to recognize these activities from intraoperative data, especially endoscopic videos~\citep{Garrow2020, Demir2023}. The recent breakthrough advances in data-driven Artificial Intelligence (AI), in particular deep learning, has fueled a surge of research activities in SDS focusing on recognizing activities from endoscopic videos; Phases~\citep{Twinanda2017, Garrow2020, Ramesh2021, Demir2023}, steps~\citep{Charriere2014, Ramesh2021, ramesh_weakly_2023}, modeling surgical actions, for example through the use of action triplets~\citep{nwoye2020recognition, nwoye2022rendezvous} and detection and localization surgical tools~\citep{AlHajj2018, Vardazaryan2018} are some of the popular \textcolor{changes}{tasks} studied in the community.

Given the data-driven nature of these recent AI methods in SDS, the availability of large labeled surgical video datasets is paramount. Datasets have been curated in the community to study phase recognition across different types of surgeries: Cholec80~\citep{Twinanda2017} for laparoscopic cholecystectomy, Bypass40~\citep{Ramesh2021} for laparoscopic Roux-en-Y gastric bypass (LRYGB), laparoscopic sleeve gastrectomy~\citep{Hashimoto2019}, transanal total mesorectal excision~\citep{Kitaguchi2021}, and laparoscopic inguinal hernia repair~\citep{Takeuchi2022}. Nevertheless, datasets to train AI models for other more fine-grained tasks, such as recognition of surgical tools, action triplets, safe dissection zones, etc, have only been collected for specific surgeries. 
For example, Bypass40~\citep{Ramesh2021} and CATARACTS\footnote{https://cataracts2020.grand-challenge.org/} have been annotated with steps alongside phases for LRYGB and cataract surgeries, CholecT50~\citep{nwoye2022rendezvous} contains surgical action triplets labels for laparoscopic cholecystectomy surgeries, \textcolor{changes}{and} safe dissection zones have been studies for cholecystectomy~\citep{Madani2020}.
Furthermore, these labeled datasets have been collected from a single medical center. Training on mono-centric datasets limits the model’s generalizability to datasets from other centers. To overcome this generalization gap multi-centric datasets representing different surgical techniques and workflows are warranted~\citep{Kitaguchi2022, mascagni_multicentric_2022, kassem_federated_2023, wagner_comparative_2023}. 
\textcolor{changes}{
One such attempt has been made in creating the HeiChole dataset which consists of 33 videos of laparoscopic cholecystectomy performed in 3 different medical centers~\citep{wagner_comparative_2023}}. However, multi-centric datasets are rare as they are difficult to acquire and annotate consistently. 

\begin{table*}[h!]
\caption{List of phases and steps of LRYGB procedure.}\label{tab:phase_step_ontology}
\centering
\small
\setlength\tabcolsep{4pt}
\begin{tabular*}{\textwidth}{lp{0.24\linewidth}lp{0.3\linewidth}lp{0.31\linewidth}}
\noalign{\smallskip}\hline\noalign{\smallskip}
\# & Phases & \# & Phases & \# & Phases \\
\noalign{\smallskip}\hline\noalign{\smallskip}
P1 & Preparation & P5 & Anastomosis test & P9 & Mesenteric defect closure\\
P2 & Gastric pouch creation & P6 &  Jejunal separation& P10 & Cleaning \& coagulation\\
P3 & Omentum division & P7 & Petersen space closure & P11 & Disassembling\\
P4 & Gastrojejunal anastomosis & P8 & Jejunojejunal anastomosis & P12 & Other intervention\\
\noalign{\smallskip}\hline\noalign{\smallskip}
\# & Steps & \# & Steps & \# & Steps\\
\noalign{\smallskip}\hline\noalign{\smallskip}
S0 & Null step & S16 & Biliary limb measurement& S32 & Jejunojejunal stapling\\
S1 & Abdominal cavity exploration & S17 & Jejunum opening& S33 & Jejunojejunal defect closure\\
S2 & Trocar placement & S18 & Gastrojejunal stapling& S34 & Jejunojejunal anastomosis reinforcement\\
S3 & Retractor placement& S19 & Gastrojejunal defect closure& S35 & Staple line reinforcement\\
S4 & Fat pad dissection& S20 & Mesenteric opening & S36& Mesenteric defect exposure\\
S5 & Lesser curvature dissection& S21 & Jejunal transection& S37 & Mesenteric defect closure\\
S6 & Horizontal stapling& S22 & Gastric tube placement& S38 & Anastomosis fixation\\
S7 & Retrogastric dissection& S23 & Jejunal clamping& S39 & Hemostasis\\
S8 & Vertical stapling& S24 & Dye injection& S40 & Irrigation/Aspiration\\
S9 & Gastric remnant reinforcement& S25 & Visual assessment & S41 & Parietal closure\\
S10 & Gastric pouch reinforcement& S26 & Gastrojejunal anastomosis reinforcement& S42 & Trocar removal\\
S11 & Gastric opening& S27 & Petersen space exposure& S43 & Suture of small bowel lesion\\
S12 & Exposure of the omentum& S28 & Petersen space closure& S44 & Drainage insertion\\
S13 & Omental transection& S29 & Biliary limb opening& S45 & Specimen retrieval\\
S14 & Adhesiolysis& S30 & \multicolumn{2}{l}{Alimentary limb measurement} & \\
S15 & Treitz angle identification& S31 & Alimentary limb opening&  & \\
\noalign{\smallskip}\hline
\end{tabular*}
\end{table*} 

Besides, only a few works have explored recognizing activities at different levels of granularity. \citep{Ramesh2021, Valderrama2022} have attempted joint phase and step recognition using endoscopic video datasets from a single medical center. 
\textcolor{changes}{The most closely related work to this paper in objectives is HeiChole~\citep{wagner_comparative_2023} which created a multi-centric dataset of 33 videos for phase recognition (7 classes), action recognition (4 classes), instrument detection (7 instruments), and skill assessment (5 classes) tasks.}
To date \textcolor{changes}{and to the best of our knowledge}, phase and step recognition have not been studied in a multi-centric dataset of endoscopic videos. 

To this end, the study has two objectives: creating a large multi-centric dataset for a complex LRYGB surgical procedure and recognition of activities at multiple levels. Thus the contributions of this work are threefold:
\begin{enumerate}
    \item Introduction of a multi-centric dataset of 140 laparoscopic Roux-en-Y gastric bypass videos from two centers (Strasbourg and Bern)
    \item \textcolor{changes}{The full annotated dataset with LRYGB ontology of 12 phases and 46 steps, which will be publicly released. The code and evaluation scripts will be made available alongside the dataset.}
    \item \textcolor{changes}{Evaluation of AI models for phase and step recognition and assessment of multi-centric model generalization}
\end{enumerate}
\section{Datasets and Annotations}

\textbf{BernBypass70} is a dataset consisting of 70 surgical videos of LRYGB at Inselspital, Bern University Hospital, Switzerland. The surgeries were performed by three surgeons. The videos were recorded at a resolution of $720\times576$ at 25 frames-per-second (fps). 
\\\\
\textbf{StrasBypass70}, extending the Bypass40~\citep{Ramesh2021} dataset, is a collection of 70 videos of LRYGB surgeries performed by surgeons at the University Hospital of Strasbourg, France. The videos were recorded at a resolution of \textcolor{changes}{$854\times480$} or \textcolor{changes}{$1920\times1080$} resolution at 25 fps and were uniformly edited to a resolution of \textcolor{changes}{$854\times480$}. 
\\\\
\textbf{MultiBypass140} is \textcolor{changes}{the} combined dataset of 140 videos from Bern and Strasbourg medical centers. Sample images of the two datasets are presented in Figure \ref{fig:sample_images}. \textcolor{changes}{All videos have been anonymized by blacking out the potentially identifying frames outside the patient's body. Those out-of-body frames have been detected using our publicly released OoBNet model ~\citep{lavanchy_preserving_2023} and were verified by manual review.}
\\\\
\textbf{Annotations.} Two board-certified surgeons with more than 10 years of clinical practice and extensive experience with surgical video analysis, annotated the MultiBypass140 dataset with activities at two levels of granularity, i.e., phases and steps. The annotation ontology of the LRYGB procedure as defined in \cite{lavanchy2023proposal} consists of 12 phases and 46 finer-grained steps, presented in Table \ref{tab:phase_step_ontology}. A detailed description of all the phases and steps can be found in the supplementary. The MultiBypass140 was annotated by two board-certified surgeons \textcolor{changes}{using the MOSaiC software~\citep{Mazellier2023Mosaic}}. The annotation inter-rater reliability (Cohen’s kappa) of the ontology between the two surgeons is found to be 96\% for phases and 81\% for steps~\citep{lavanchy2023proposal}. 
\\\\
\textbf{Data Statistics.} On average, the surgical duration is 110  and 72 minutes and the total number of frames at 1 fps amounts to 464,794 and 305,907 in the StrasBypass70 and BernBypass70, respectively.  
Data characteristics of the multi-center dataset can be found in the supplementary. 
\begin{figure*}[t!]
     \centering
     \includegraphics[clip, trim=0.cm 4.cm 1.5cm 1.0cm, width=0.9\linewidth]{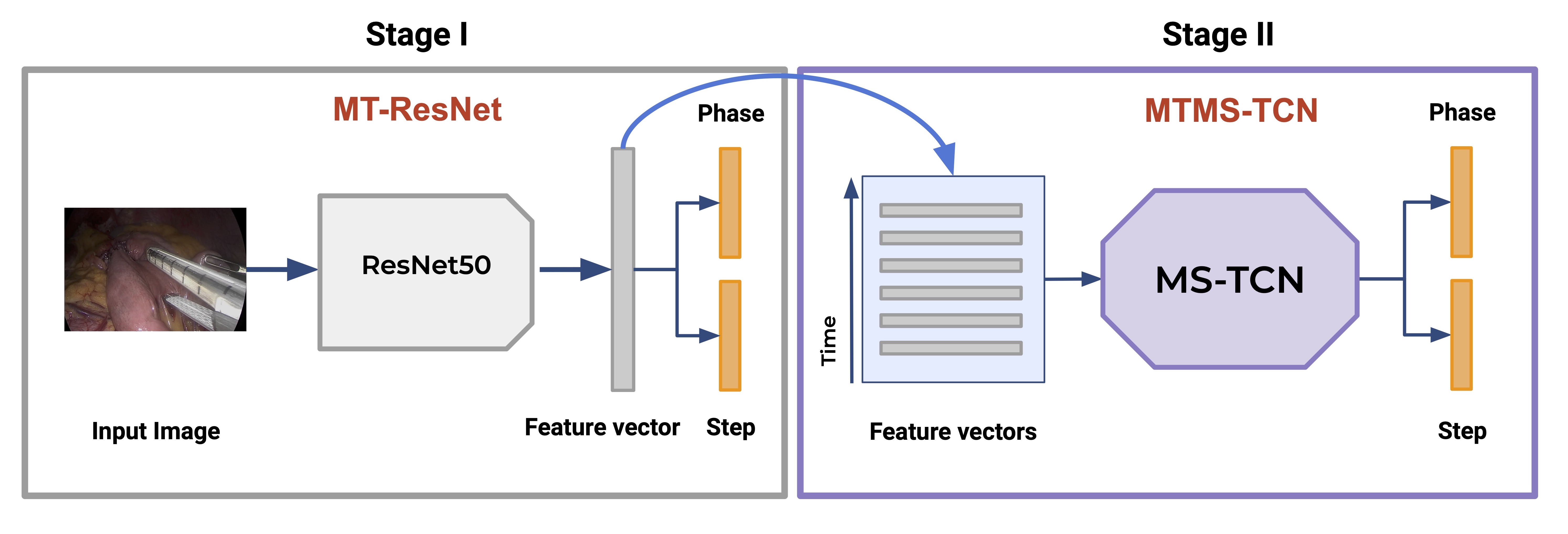}
     \caption{Schematic representation of the model architecture. In stage I the input images are processed by a ResNet-50 convolutional neural network to extract visual feature vectors, that are passed to stage II. In stage II the feature vectors of subsequent images of a video are stacked and processed by a multi-stage temporal convolutional network (MS-TCN). This introduces temporal awareness into the model predictions.} \label{fig:method}
\end{figure*}
According to video duration, StrasBypass70 and BernBypass70 were split into training (40 videos each), validation (10 videos each), and test set (20 videos each), resulting in 80 training, 20 validation, and 40 test videos for MultiBypass140.

\subsection{Model architecture}

A state-of-the-art deep learning model, MTMS-TCN~\citep{Ramesh2021}, for surgical activity recognition, was used for the different experiments presented in this paper. The pipeline of MTMS-TCN consists of two stages where first a multi-task Convolutional Neural Network (CNN) (ResNet-50~\citep{He2016}) model is employed for extracting visual features from images followed by a multi-task multi-stage causal Temporal Convolutional Network (TCN) to refine the features and extracting temporal information for joint phase and step recognition. A schematic representation of the model architecture is displayed in Figure~\ref{fig:method}. 
\\ \\
\textbf{Spatial model:} ResNet-50~\citep{He2016} is one of the popular CNN architectures that has been heavily utilized in the computer vision community for activity recognition. Due to its success, ResNet-50 is utilized as a visual feature extractor and trained in multi-task learning of phase and step recognition. The model is initialized with pre-trained weights on ImageNet (a public dataset of over 14 million images for object recognition tasks) and trained using Adam optimizer for 30 epochs. 
\\ \\
\textbf{Temporal model:} MTMS-TCN, is a two-stage TCN model that was trained in a multi-task learning setup on video features extracted from the CNN model for 200 epochs. Furthermore, each stage of the TCN model consists of causal convolution\textcolor{changes}{s} that utilize only information from past frames and dilated convolutions with exponentially increasing dilation factor that facilitates capturing long temporal dependencies. This work utilized MTMS-TCN with a single-stage temporal model as a second stage does not improve the model performance~\citep{Ramesh2021}. 

\subsection{Experiments}
\begin{figure}[t!]
     \centering
     \includegraphics[width=0.50\textwidth, trim={0cm {.07\textwidth} 0cm {.05\textwidth}}, clip]{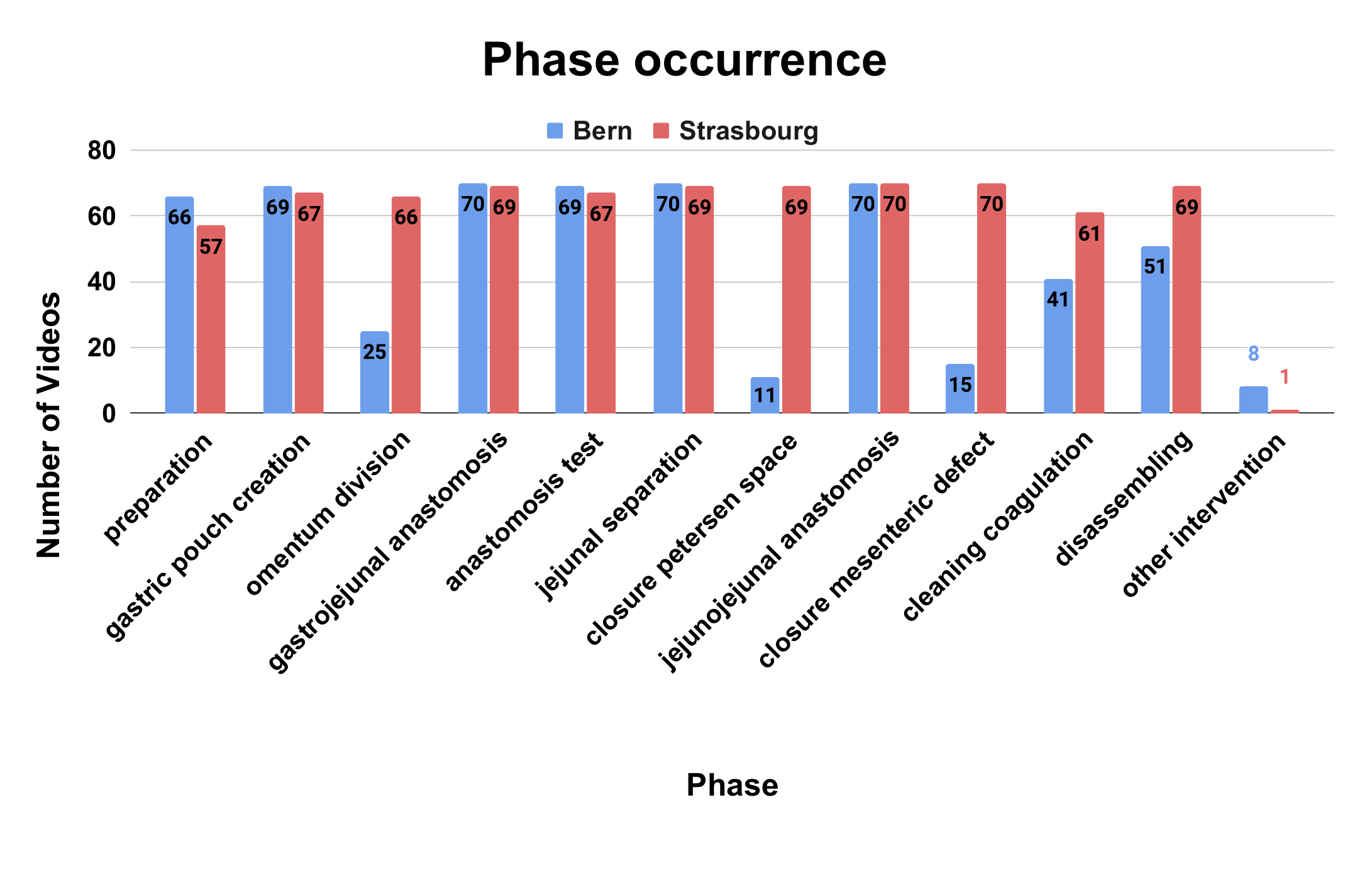}
     \caption{Total occurrence of phases in the videos from the two medical centers.}
    \label{fig:phase_occurence_duration}
\end{figure}

\begin{figure*}[!ht]
     \centering
     \includegraphics[angle=270, width=0.74\textwidth, trim={0cm {.07\textwidth} 0cm {.1\textwidth}}, clip]{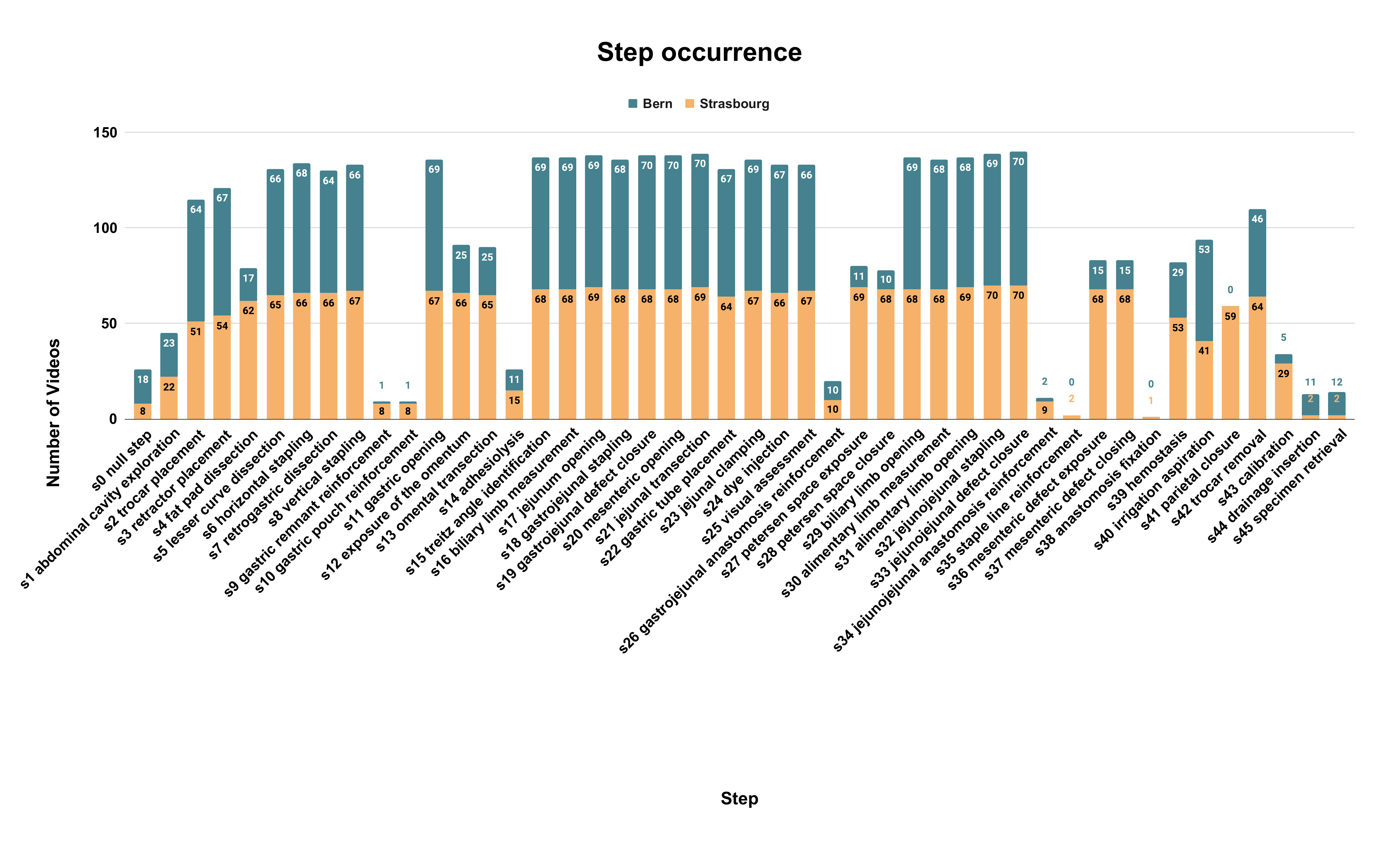}
     \caption{Total occurrence of steps in the videos from the two medical centers.}
    \label{fig:steps_occurence_duration}
\end{figure*}
To establish benchmarks for phase and step recognition on BernBypass70, StrasBypass70, and on the joint MultiBypass140 dataset 5 different model architectures were assessed: 1) A ResNet-50 (CNN)~\citep{He2016}, 2) a long short-term memory (LSTM)~\citep{hochreiter_LSTM_1996}, 3) Multi-task LSTM (MT-LSTM), 4) a multi-stage TCN (TeCNO)~\citep{Czempiel2020TeCNOSP}, and 5) a MTMS-TCN ~\citep{Ramesh2021}.

Seven experimental setups were used to analyze the generalizability of MTMS-TCN: 
\begin{enumerate}
    \item Training and evaluation on BernBypass70 
    \item Training and evaluation on StrasBypass70 
    \item Training and evaluation on the joint MultiBypass140 dataset 
    \item Training on BernBypass70 and evaluation on StrasBypass70 
    \item Training on StrasBypass70 and evaluation on BernBypass70 
    \item Training on MultiBypass140 and evaluat\textcolor{changes}{ion} on BernBypass70 
    \item Training on MultiBypass140 and evaluation on StrasBypass70 
\end{enumerate}

\subsection{Model evaluation}

Model performance was assessed by comparing human ground truth annotations with model predictions measuring accuracy, precision, recall, and F1-score. 
Accuracy is the proportion of all correct (true positive and true negative) predictions among all predictions.
Precision\textcolor{changes}{,} also referred to as positive predictive value, is the proportion of true positive predictions among all positive (true and false positive) predictions. Recall, also referred to as sensitivity, is the proportion of true positive predictions among all relevant (true positive and false negative) predictions. F1-score, which is a measure of accuracy, is defined as the harmonic mean of precision and recall. As in previous works, performance metrics were averaged across phases and steps per video and then across videos~\citep{Ramesh2021, Czempiel2020TeCNOSP}.

\section{Results \& Discussions}

\begin{table*}
\caption{Benchmark of phase recognition. (Best results are in bold)}~\label{tab:phase_table}
\centering
\setlength\tabcolsep{5pt}
\begin{tabular*}{0.8\textwidth}{llcccc}
\hline\noalign{\smallskip}
Dataset & Model & ACC (\%) & PR (\%) & RE (\%) & F1 (\%)\\
\noalign{\smallskip}\hline\noalign{\smallskip}
& CNN & $ 74.53 \pm 13.34 $ & $ 44.79 \pm 8.44 $ & $ 45.69 \pm 8.19 $ & $ 42.38 \pm 9.14 $ \\
& LSTM & $ 79.73 \pm 13.75 $ & $ 54.91 \pm 9.31 $ & $ 56.19 \pm 10.24 $ & $ 52.60 \pm 10.34 $ \\
1) BernBypass70 & MT-LSTM &$ 80.69 \pm 13.85 $ & $ 56.98 \pm 11.54 $ & $ 57.14 \pm 13.38 $ & $ 54.15 \pm 12.84 $ \\
& TeCNO & $ 83.81 \pm 13.55 $ & $ 61.28 \pm 13.84 $ & $ 62.81 \pm 14.07 $ & $ 59.22 \pm 14.56 $ \\
& MTMS-TCN & $ \textbf{85.30} \pm \textbf{13.19} $ & $ \textbf{64.62} \pm \textbf{11.33} $ & $ \textbf{67.41} \pm \textbf{13.81} $ & $ \textbf{62.40} \pm \textbf{12.87} $\\
\noalign{\smallskip}\hline\noalign{\smallskip}
& CNN & $ 82.46 \pm 7.90 $ & $ 72.91 \pm 9.17 $ & $ 73.37 \pm 8.67 $ & $ 71.13 \pm 9.47 $ \\
& LSTM & $ 86.37 \pm 7.68 $ & $ 76.66 \pm 9.52 $ & $ 80.90 \pm 9.63 $ & $ 76.42 \pm 10.35 $ \\
2) StrasBypass70 & MT-LSTM & $ 86.16 \pm 8.61 $ & $ 79.87 \pm 9.31 $ & $ 79.16 \pm 8.94 $ & $ 77.45 \pm 10.06 $ \\
& TeCNO & $ 89.50 \pm 7.55 $ & $ \textbf{81.17} \pm \textbf{8.54} $ & $ \textbf{84.26} \pm \textbf{7.73} $ & $ \textbf{80.70} \pm \textbf{8.81} $ \\
& MTMS-TCN & $ \textbf{90.23} \pm \textbf{7.04} $ & $ 80.48 \pm 9.37 $ & $ 82.39 \pm 8.22 $ & $ 79.87 \pm 9.37 $ \\
\noalign{\smallskip}\hline\noalign{\smallskip}
& CNN & $ 78.18 \pm 11.21 $ & $ 57.43 \pm 15.87 $ & $ 56.85 \pm 15.36 $ & $ 54.8 \pm 15.63 $ \\
& LSTM & $ 82.56 \pm 11.89 $ & $ 68.18 \pm 14.11 $ & $ 68.15 \pm 13.8 $ & $ 65.02 \pm 14.22 $ \\
3) MultiBypass140 & MT-LSTM & $ 83.94 \pm 11.18 $ & $ 67.58 \pm 14.93 $ & $ 66.88 \pm 15.67 $ & $ 64.86 \pm 15.97 $ \\
& TeCNO & $ 86.44 \pm 10.77 $ & $ \textbf{72.59} \pm \textbf{13.99} $ & $ \textbf{75.3} \pm \textbf{12.35} $ & $ 71.03 \pm 14.02 $ \\
& MTMS-TCN & $ \textbf{87.91} \pm \textbf{10.64} $ & $ 72.27 \pm 13.13 $ & $ 74.82 \pm 13.36 $ & $ \textbf{71.28} \pm \textbf{13.96} $ \\
\noalign{\smallskip}\hline
\end{tabular*}
\end{table*}
\begin{table*}
\caption{Benchmark of step recognition. (Best results are in bold)}~\label{tab:step_table}
\centering
\setlength\tabcolsep{5pt}
\begin{tabular*}{0.8\textwidth}{llcccc}
\hline\noalign{\smallskip}
Dataset & Model & ACC (\%) & PR (\%) & RE (\%) & F1 (\%)\\
\noalign{\smallskip}\hline\noalign{\smallskip}
& CNN & $ 58.92 \pm 11.63 $ & $ 38.26 \pm 7.95 $ & $ 38.47 \pm 7.39 $ & $ 35.55 \pm 7.44 $ \\
& LSTM & $ 64.99 \pm 12.44 $ & $ 48.66 \pm 10.91 $ & $ 48.66 \pm 11.12 $ & $ 44.88 \pm 10.53 $ \\
1) BernBypass70 & MT-LSTM & $ 63.54 \pm 13.92 $ & $ 49.40 \pm 11.21 $ & $ 48.49 \pm 12.37 $ & $ 44.93 \pm 11.48 $ \\
& TeCNO & $ 67.54 \pm 13.49 $ & $ 50.47 \pm 10.42 $ & $ \textbf{53.01} \pm \textbf{11.74} $ & $ 47.56 \pm 10.85 $ \\
& MTMS-TCN & $ \textbf{67.54} \pm \textbf{13.28} $ & $ \textbf{51.04} \pm \textbf{10.36} $ & $ 52.84 \pm 10.44 $ & $ \textbf{47.99} \pm \textbf{10.23} $ \\
\noalign{\smallskip}\hline\noalign{\smallskip}
& CNN & $ 70.44 \pm 11.48 $ & $ 50.29 \pm 7.1 $ & $ 50.66 \pm 8.4 $ & $ 47.67 \pm 8.19 $ \\
& LSTM & $ 75.26 \pm 11.67 $ & $ 60.15 \pm 7.35 $ & $ 58.74 \pm 9.04 $ & $ 56.37 \pm 9.05 $ \\
2) StrasBypass70 & MT-LSTM & $ 74.67 \pm 11.48 $ & $ 58.98 \pm 8.10 $ & $ 59.27 \pm 9.73 $ & $ 56.10 \pm 9.33 $ \\
& TeCNO & $ \textbf{78.49} \pm \textbf{9.43} $ & $ \textbf{60.15} \pm \textbf{6.92} $ & $ \textbf{62.09} \pm \textbf{8.11} $ & $ \textbf{58.13} \pm \textbf{7.87} $ \\
& MTMS-TCN & $ 77.78 \pm 10.24 $ & $ 59.14 \pm 7.84 $ & $ 61.28 \pm 8.65 $ & $ 57.27 \pm 8.47 $ \\
\noalign{\smallskip}\hline\noalign{\smallskip}
& CNN & $ 65.21 \pm 12.75 $ & $ 44.19 \pm 10.07 $ & $ 44.47 \pm 10.55 $ & $ 41.47 \pm 10.31 $ \\
& LSTM & $ 70.18 \pm 13.04 $ & $ 54.74 \pm 11.71 $ & $ 54.24 \pm 12.55 $ & $ 51.15 \pm 12.35 $ \\
3) MultiBypass140 & MT-LSTM & $ 69.55 \pm 13.76 $ & $ 53.92 \pm 11.64 $ & $ 53.14 \pm 12.64 $ & $ 50.11 \pm 12.45 $ \\
& TeCNO & $ \textbf{73.49} \pm \textbf{13.17} $ & $ \textbf{55.81} \pm \textbf{11.1} $ & $ \textbf{57.29} \pm \textbf{12.18} $ & $ \textbf{53.08} \pm \textbf{11.95} $ \\
& MTMS-TCN & $ 72.85 \pm 12.68 $ & $ 55.32 \pm 10.55 $ & $ 56.58 \pm 11.7 $ & $ 52.59 \pm 11.32 $ \\
\noalign{\smallskip}\hline
\end{tabular*}
\end{table*}

This is the first study to evaluate deep learning models for multi-level activity recognition, i.e., phases and steps, on a large multi-centric video dataset of LRYGB procedures. A common ontology of phases and steps has been designed to capture the surgical workflow followed in the two medical centers. In this section, we present the results on the new multi-centric video dataset of LRYGB procedures and highlight our findings.
\\\\
\textbf{Workflow: Strasbourg vs Bern.} Differences in surgical workflow between medical centers are natural and common, as different surgeons perform the interventions. StrasBypass70 on average has a video duration of 111$\pm$33 minutes and the average workflow includes 10 phases and 33 steps. On the other hand, BernBypass70 has an average video duration of 73$\pm$20 minutes with the average workflow consisting of 8 phases and 27 steps.  To understand the surgical workflow differences of LRYGB followed in the Strasbourg and Bern medical centers, we visualize and compare the phase and step occurrences in Figure \ref{fig:phase_occurence_duration} \& \ref{fig:steps_occurence_duration}. We also visualize the surgical workflows, modeled as phase transition graphs, in Figure \ref{fig:sample_images}. 

In StrasBypass70, the occurrence of phases and steps is evenly distributed. Either a phase or a step occurs in most videos in the dataset, or it does not occur at all. In contrast, BernBypass70 has only some videos containing all phases and steps. Most of the videos contain a subset of phases and steps. These differences in dataset distribution of phases and steps between StrasBypass70 and BernBypass70 result from differences in surgical technique and workflows. The phase transition graph visualized in Figure \ref{fig:sample_images} further exemplifies these differences in surgical workflows across different centers. 
In StrasBypass70 the omentum is routinely divided (P3) and both mesenteric defects are routinely closed (P7 \& P9), which is not routinely done in BernBypass70. Given the hierarchical structure of phases and steps, with every phase missing, corresponding steps are missing as well. Hence, the average video of BernBypass70 contains 2 phases and 6 steps less than the average StrasBypass70 video. This finding is also reflected by the average video duration which is 38 minutes shorter in BernBypass70 compared to StrasBypass70 videos. 
\\\\
\textbf{\textcolor{changes}{Recognition:} Individual centers.} To independently analyze the performance of deep learning models on each center/dataset, we train different models on BernBypass70, StrasBypass70, and MultiBypass140 datasets and evaluate the models' performance on respective test sets. The phase and step recognition task results are presented in Table \ref{tab:phase_table} \& \ref{tab:step_table}. 

\begin{table*}
\caption{Cross dataset evaluation of MTMS-TCN.}~\label{tab:cross_eval_table}
\centering
\setlength\tabcolsep{5pt}
\begin{tabular*}{0.83\textwidth}{llcccc}
\hline\noalign{\smallskip}
Experiment & Model & ACC (\%) & PR (\%) & RE (\%) & F1 (\%)\\
\noalign{\smallskip}\hline\noalign{\smallskip}
\multicolumn{6}{c}{Phase} \\
\noalign{\smallskip}
4) BernBypass70 $\rightarrow$ & CNN & $ 57.34 \pm 8.52 $ & $ 35.94 \pm 6.16 $ & $ 45.41 \pm 6.51 $ & $ 32.72 \pm 5.47 $\\
StrasBypass70 & MTMS-TCN & $ 64.44 \pm 7.91 $ & $ 36.76 \pm 5.49 $ & $ 40.16 \pm 7.38 $ & $ 33.10 \pm 5.72 $\\
\noalign{\smallskip}\hdashline\noalign{\smallskip}
5) StrasBypass70 $\rightarrow$ & CNN & $ 56.66 \pm 14.48 $ & $ 32.14 \pm 7.61 $ & $ 34.13 \pm 7.36 $ & $ 29.54 \pm 8.21 $ \\
BernBypass70 & MTMS-TCN & $ 72.36 \pm 17.57 $ & $ 42.21 \pm 9.80 $ & $ 45.13 \pm 13.55 $ & $ 39.05 \pm 11.95 $ \\
\noalign{\smallskip}\hdashline\noalign{\smallskip}
6) MultiBypass140 $\rightarrow$ & CNN & $ 76.77 \pm 12.34 $ & $ 46.48 \pm 7.41 $ & $ 46.90 \pm 8.72 $ & $ 43.99 \pm 8.29 $ \\
BernBypass70 & MTMS-TCN & $ 85.62 \pm 12.74 $ & $ 62.13 \pm 8.34 $ & $ 65.02 \pm 10.56 $ & $ 60.63 \pm 9.49 $ \\
\noalign{\smallskip}\hdashline\noalign{\smallskip}
7) MultiBypass140 $\rightarrow$ & CNN & $ 83.30 \pm 8.03 $ & $ 70.85 \pm 8.18 $ & $ 71.70 \pm 8.36 $ & $ 69.46 \pm 8.75 $ \\
StrasBypass70 & MTMS-TCN & $ 90.19 \pm 7.31 $ & $ 82.41 \pm 8.33 $ & $ 84.63 \pm 7.31 $ & $ 81.93 \pm 8.54 $ \\
\noalign{\smallskip}\hline\noalign{\smallskip}
\multicolumn{6}{c}{Step} \\
\noalign{\smallskip}
4) BernBypass70 $\rightarrow$ & CNN & $ 40.16 \pm 9.65 $ & $ 26.12 \pm 4.55 $ & $ 27.82 \pm 5.65 $ & $ 20.99 \pm 4.36 $ \\
StrasBypass70 & MTMS-TCN & $ 44.87 \pm 10.42 $ & $ 29.05 \pm 5.96 $ & $ 29.16 \pm 5.59 $ & $ 23.81 \pm 5.63 $\\
\noalign{\smallskip}\hdashline\noalign{\smallskip}
5) StrasBypass70 $\rightarrow$ & CNN & $ 37.45 \pm 11.48 $ & $ 18.51 \pm 4.74 $ & $ 21.41 \pm 3.78 $ & $ 17.35 \pm 4.56 $ \\
BernBypass70 & MTMS-TCN & $ 49.00 \pm 15.14 $ & $ 24.98 \pm 6.52 $ & $ 29.01 \pm 7.74 $ & $ 23.23 \pm 6.56 $ \\
\noalign{\smallskip}\hdashline\noalign{\smallskip}
6) MultiBypass140 $\rightarrow$ & CNN & $ 57.19 \pm 12.07 $ & $ 36.18 \pm 7.29 $ & $ 36.09 \pm 7.53 $ & $ 33.25 \pm 7.42 $ \\
BernBypass70 & MTMS-TCN & $ 67.74 \pm 13.05 $ & $ 50.06 \pm 10.99 $ & $ 51.06 \pm 12.34 $ & $ 46.82 \pm 11.35 $ \\
\noalign{\smallskip}\hdashline\noalign{\smallskip}
7) MultiBypass140 $\rightarrow$ & CNN &  $ 70.23 \pm 11.36 $ & $ 50.33 \pm 6.87 $ & $ 50.49 \pm 7.54 $ & $ 47.45 \pm 7.73 $ \\
StrasBypass70 & MTMS-TCN & $ 77.96 \pm 9.96 $ & $ 60.59 \pm 6.83 $ & $ 62.11 \pm 7.78 $ & $ 58.35 \pm 7.80 $ \\
\noalign{\smallskip}\hline
\end{tabular*}
\end{table*}

\begin{figure*}[t!]
    \centering
    \begin{subfigure}[b]{0.9\textwidth}
        \includegraphics[width=\textwidth]{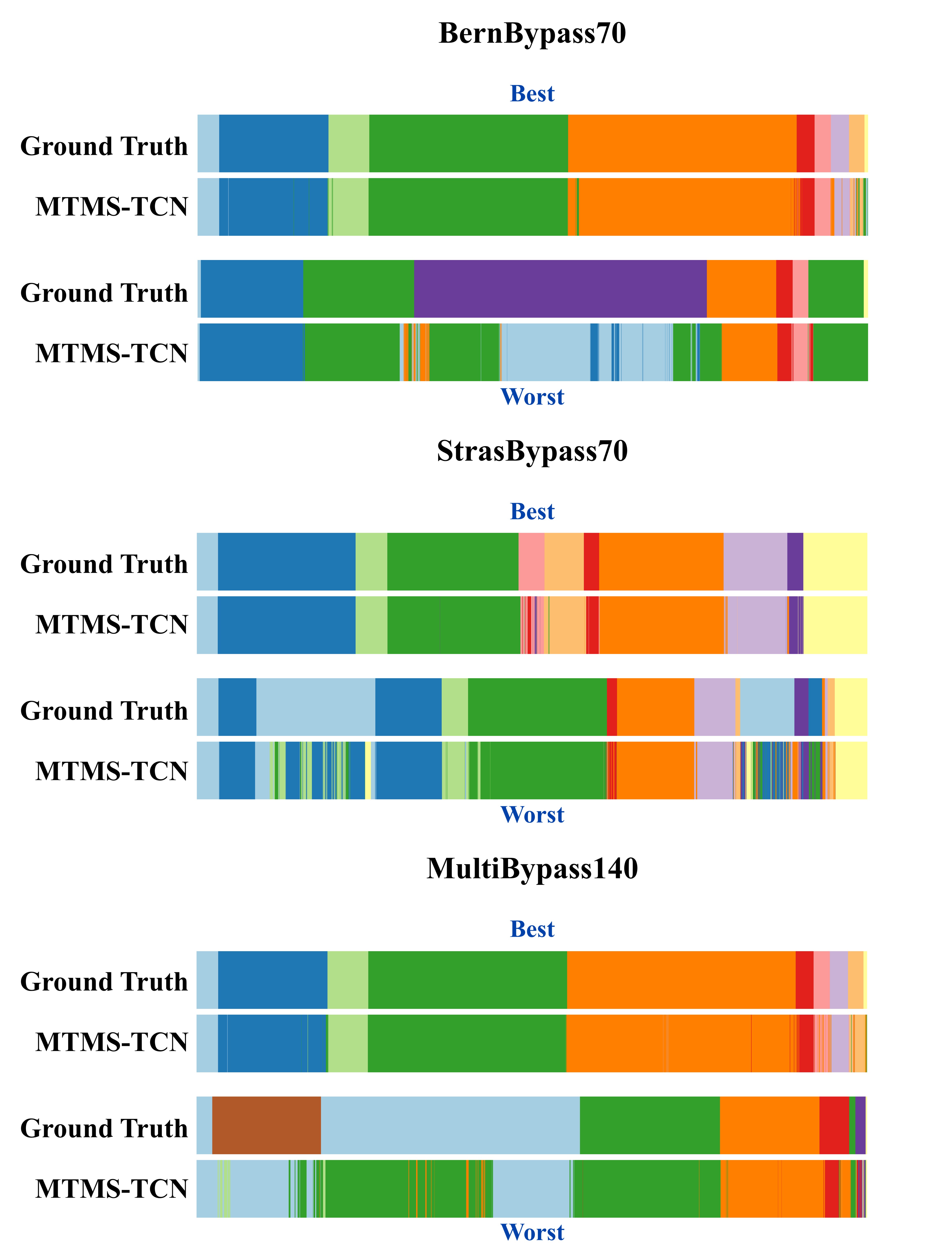}
    \end{subfigure}
    \par\smallskip
    \begin{subfigure}[b]{0.8\textwidth}
        \includegraphics[width=\textwidth]{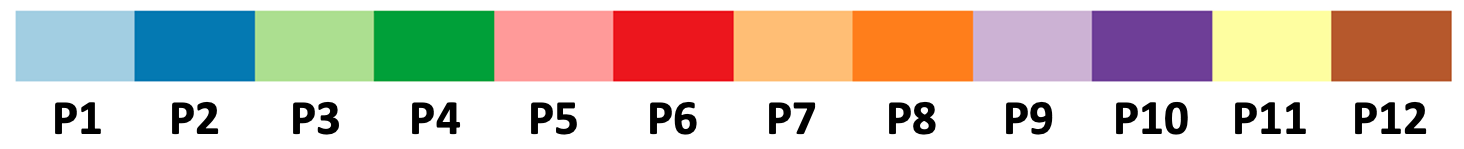}
    \end{subfigure}
    \caption{Best (upper row) and worst (lower row) pairs of ground truth annotations (top) and model predictions (MTMS-TCN, bottom) for all 3 datasets. Every pair corresponds to one video and the width of each phase is relative to its duration.}
    \label{fig:phase_predictions}
\end{figure*}

\begin{figure*}[t!]
\centering
    \begin{subfigure}[b]{0.85\textwidth}
        \includegraphics[width=\textwidth]{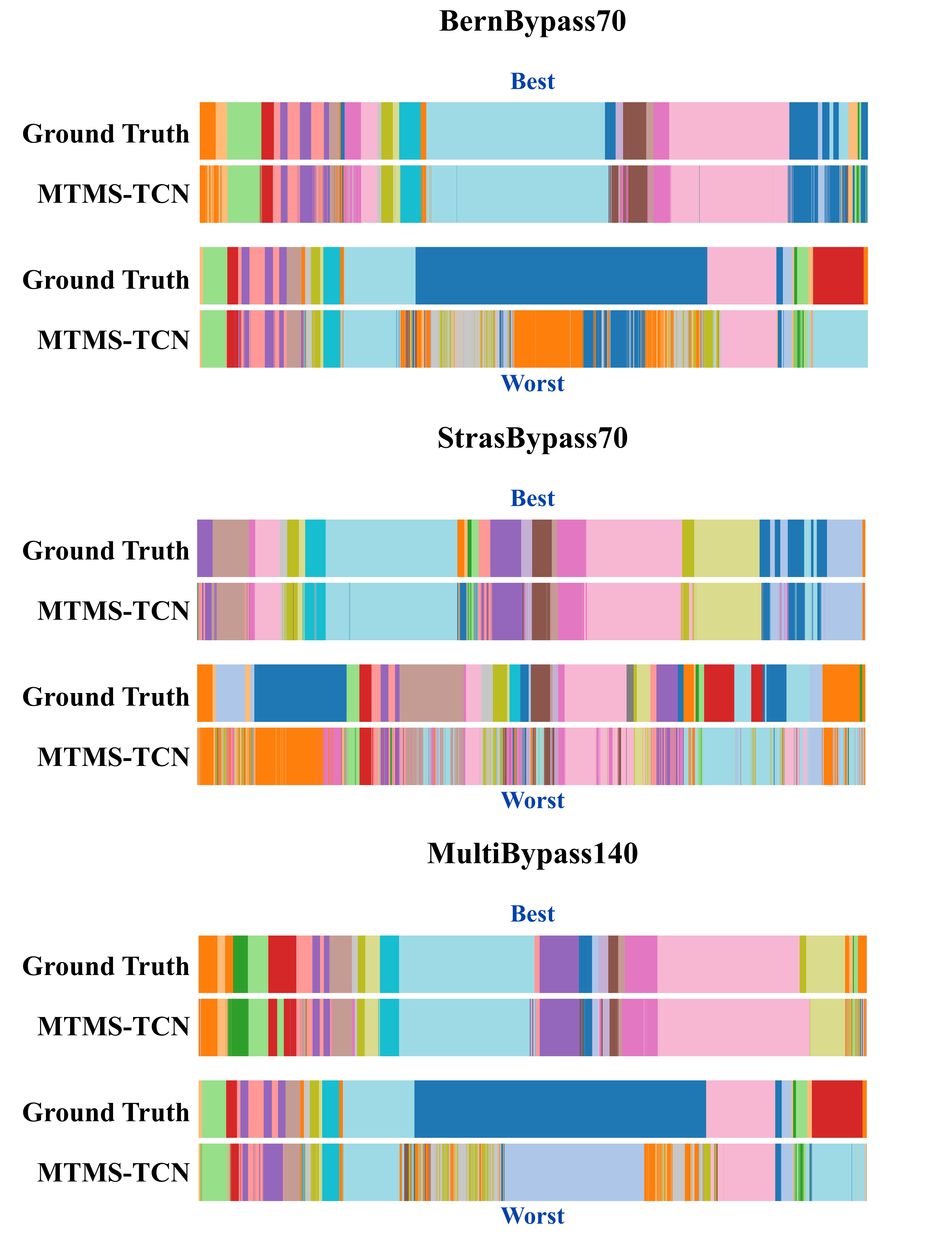}
    \end{subfigure}
    \par\smallskip
    \begin{subfigure}[b]{0.85\textwidth}
        \centering
        \includegraphics[width=\textwidth]{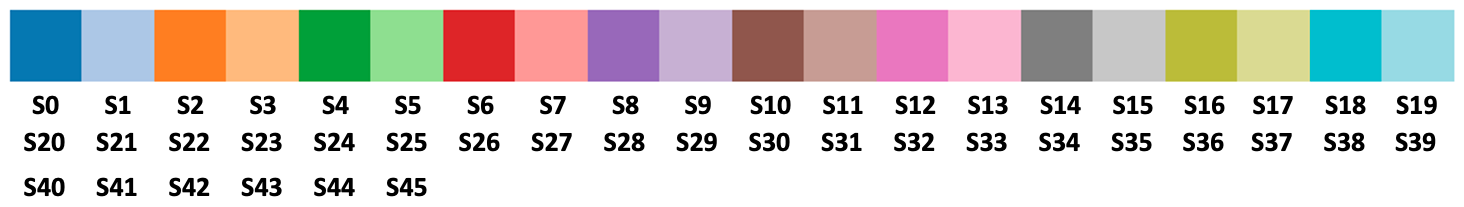}
    \end{subfigure}
    \caption{Best (upper row) and worst (lower row) pairs of ground truth annotations (top) and model predictions (MTMS-TCN, bottom) for all 3 datasets. Every pair corresponds to one video and the width of each step is relative to its duration.}
    \label{fig:step_predictions}
\end{figure*}

All the models, both spatial and spatio-temporal, achieve considerably low performance across all the metrics on BernBypass70 in comparison to StrasBypass70. For instance, the CNN (ResNet-50) spatial model on phase recognition task shows 8\% lower accuracy and a staggering 28\% degradation in F1-score on BernBypass70 compared to StrasBypass70. Spatio-temporal model, MTMS-TCN, performs 5\% lower in accuracy and 15-17\% lower on all other metrics on BernBypass70 over StrasBypass70. Similarly for step recognition, CNN and MTMS-TCN on BernBypass70 achieve 12\% and 8-10\% lower than StrasBypass70 on all metrics. These differences are direct consequences of the differences in surgical workflow followed in the two centers \textcolor{changes}{and consistent with previous work on laparoscopic cholecystectomy~\citep{kassem_federated_2023}}. Given that many phases and steps (Figure \ref{fig:phase_occurence_duration} \& \ref{fig:steps_occurence_duration}) are not carried out routinely in Bern compared to Strasbourg medical center, their occurrences/class distribution is notably skewed in BernBypass70 which makes recognition of phases and steps increasingly challenging for deep learning models on this dataset. \textcolor{changes}{This can be witnessed in Figure \ref{fig:phase_predictions} \& \ref{fig:step_predictions} where the model performs best on videos following common workflow (P1$\rightarrow$P2$\rightarrow$P3$\rightarrow$...) in both the datasets while performs worse when there is unexpected flow of phases/steps performed during surgeries (P4$\rightarrow$P10$\rightarrow$P8 or P1$\rightarrow$P12$\rightarrow$P1$\rightarrow$P4).}

Lastly, all the deep learning models on the combined MultiBypass140 dataset have a performance \textcolor{changes}{exceeding} the performance \textcolor{changes}{on BernBypass70, but inferior to the performance on StrasBypass70.}
\\\\
\textbf{\textcolor{changes}{Recognition:} Cross-center.} \textcolor{changes}{Here we} examine the ability of the models to transfer knowledge learnt using data from one center to the other, we train CNN and MTMS-TCN on one center and evaluate them on the other (experiments 4, 5, 6, \& 7). The experimental results are tabulated in Table \ref{tab:cross_eval_table}.

The performance of the CNN \& MTMS-TCN in these experiments is considerably inferior to training and evaluation on individual mono-centric datasets (experiments 1 \& 2). CNN \& MTMS-TCN trained on BernBypass70 when evaluated on StrasBypass70 without any fine-tuning achieve 57\% \& 64\% in accuracy and 32\% \& 33\% in F1 score. This is due to the significant differences in the workflow followed in the Bern center with many phases and steps not routinely carried out. 
Inversely, CNN \& MTMS-TCN achieves 56\% \& 72\% in accuracy and 29\% \& 39\% in F1 when trained on StrasBypass70 and evaluated on BernBypass70. Although in StrasBypass70 the occurrence of phases and steps are evenly distributed, the knowledge learned by both spatial and spatio-temporal models (CNN \& MTMS-TCN) on StrasBypass70 is still not transferable to BernBypass70. This odd performance could be for two reasons: 1) The variability in visual appearance between centers caused due to different instruments, lighting, patients' age, height, weight, gender, and ethnicity affects the performance of CNN\textcolor{changes}{;} 2) Alongside this, the temporal differences caused due to changes in the surgical workflow across surgeons and medical centers affect the performance of MTMS-TCN. 

Both CNN \& MTMS-TCN trained on MultiBypass140 when evaluated on the mono-centric datasets (experiments 6 \& 7) achieve performance close to its performance when trained and evaluated on the individual dataset (experiments 1 \& 2) for both the phase and step recognition tasks. This shows the capacity of deep learning models to learn all the variations existing in the data and domain without any degradation in performance.
\\\\
\textbf{Challenges.} Despite its multi-centric design, this study is limited by the fact that datasets from only two centers are involved. The significant variability in surgical technique and image domain makes the transferability of deep learning models between centers a challenging task. More studies on adding video datasets from other clinical centers are imperative to capture the variability in surgical technique and dataset distributions. Furthermore, these studies could facilitate standardizing surgical procedures across the globe. 
However, labeling large surgical datasets from different centers is difficult and time-consuming. Future studies should focus on developing deep learning models to learn from a large corpus of unlabeled data from multiple centers and evaluate their performance on a multi-centric dataset. 

\section{Conclusion}

This study demonstrates the need to exhibit the variation of surgical techniques and workflow to deep learning models to avoid the generalization gap described in the literature~\citep{Kitaguchi2022, Bar2020}. With extensive experiments, the origin of the performance differences in our datasets has been investigated. It has been shown that dataset distribution and size due to different LRYGB techniques and workflows between centers have a major impact on model performance. This work highlights the importance of multi-centric datasets for the training and evaluation of AI models in surgical video analysis. The public release of the datasets and code of this work will inspire and foster future research in developing multi-centric generalizable AI models for the recognition of surgical activities.
\subsubsection*{Acknowledgements} This study was funded by the Swiss National Science Foundation (Joël L. Lavanchy: P500PM\_206724, P5R5PM\_217663), the European Union’s Horizon 2020 research and innovation program under the Marie Sklodowska-Curie grant agreement No 813782 – project ATLAS (Sanat Ramesh) and French state funds managed by the ANR within the National AI Chair program under Grant ANR-20-CHIA-0029-01 (Nicolas Padoy) and the Investments for the future program under Grant ANR-10-IAHU-02 (IHU Strasbourg). \textcolor{changes} {This work was also supported by French state funds managed within the Investissements d’Avenir program by BPI France (project CONDOR).}  Access to the HPC resources of IDRIS was granted under the allocations AD011012832R1.

\bibliographystyle{model2-names.bst}\biboptions{authoryear}
\bibliography{reference}


\clearpage
\newpage
\renewcommand{\thesubsection}{\roman{subsection}}
\appendix
\clearpage
\onecolumn
\setcounter{page}{1}

\section*{\centerline{\large Limitations in Multi-centric Generalization: Phase and Step Recognition in LRYGB Surgery}}
\subsection*{\centerline{===== ~~~Supplementary Material~~~ =====}}
\begin{center}
    \rule{\linewidth}{0.88pt}
\end{center}

\section{Phase and Step Definitions}

All the 12 phases and their definitions have been tabulated in Table \ref{tab1:phase_definition}. Similarly, all the 46 steps of a laparoscopic Roux-en-Y gastric bypass, along with their definitions, are presented in Table \ref{tab1:step_definition}.

\setlength{\tabcolsep}{6pt}
\begin{table}[h!]
\centering
\caption{Definitions of the 12 phases of laparoscopic Roux-en-Y gastric bypass surgery.}\label{tab1:phase_definition}
\begin{tabular}{p{0.05\linewidth} p{0.2\linewidth} p{0.65\linewidth}}
\hline\noalign{\smallskip}
Phase ID    &   Phase Name & Description\\ 
\noalign{\smallskip}\hline\noalign{\smallskip}
P1 &  Preparation & Access to the abdominal cavity, installation of the ports and exposure of the operating field \\
P2 & Gastric pouch creation & The proximal part of the stomach is separated from the rest to create a gastric pouch \\
P3 &  Omentum division & Vertical transection of the omentum majus to facilitate the ascent of the small bowel to the gastric pouch\\
P4 & Gastrojejunal anastomosis & Anastomosis of the small bowel with the gastric pouch \\
P5 & Anastomosis test & Verification that the gastrojejunostomy does not leak \\
P6 &  Jejunal separation & Separation of the proximal alimentary limb and the biliary limb by transection of the jejunum\\
P7 &  Petersen space closure & Closure of the Petersen space between the alimentary limb and the transverse mesocolon\\
P8 & Jejunojejunal anastomosis & Anastomosis of the distal alimentary limb with the biliary limb\\
P9 &  Mesenteric defect closure &  Closure of the mesenteric defect at the jejunojejunostomy \\
P10 &  Cleaning \& coagulation & Irrigation and aspiration of liquid/blood in the abdominal cavity, hemostasis\\
P11 &  Disassembling & Removal of the surgical instruments, retractor, ports, and camera \\
P12 & Other intervention & If any additional intervention is performed (e.g. liver biopsy, cholecystectomy) \\
\noalign{\smallskip}\hline
\end{tabular}
\end{table}

\setlength{\tabcolsep}{6pt}
\begin{longtable}{p{0.05\linewidth} p{0.2\linewidth} p{0.65\linewidth}}
\caption{Definitions of the 46 steps of laparoscopic Roux-en-Y gastric bypass surgery.}\label{tab1:step_definition}\\
\hline\noalign{\smallskip}
Step ID    &   Step Name & Description\\ 
\noalign{\smallskip}\hline\noalign{\smallskip}
\endfirsthead
\multicolumn{3}{c}
{\tablename\ \thetable\ -- \textit{Continued from previous page}}\\
\hline\noalign{\smallskip}
Step ID & Step Name & Description\\ 
\noalign{\smallskip}\hline\noalign{\smallskip}
\endhead
\hline\multicolumn{3}{r}{\textit{Continued on next page}}
\endfoot
\endlastfoot
S0 &  Null step & The camera is static and no actions are performed by the surgeon\\
S1 &  Abdominal cavity exploration &  The abdominal cavity is explored to detect alterations that could modify or prevent the planned surgery\\
S2 &  Trocar placement & Accessory trocars are introduced into the abdominal cavity\\
S3 &  Retractor placement &  Introduction and placement of a liver retractor to expose the esophagogastric junction \\
S4 & Fat pad dissection &  Dissection of the fatty tissue surrounding the esophagogastric junction to expose the angle of his and remove adhesions to the spleen \\
S5 & Lesser curvature dissection &  Opening of a retrogastric window at the lesser curvature of the stomach to facilitate the passage of the stapler\\
S6 & Horizontal stapling & Horizontal transection of the stomach with the stapler starting from the lesser curvature to create the horizontal part of the pouch\\
S7 & Retrogastric dissection & Dissection of the tissue dorsal to the stomach for better exposure\\
S8 & Vertical stapling & Vertical transection of the stomach with the stapler to create the vertical portion of the pouch\\
S9 &  Gastric remnant reinforcement & Reinforcement of the gastric remnant staple line with a suture\\
S10 &  Gastric pouch reinforcement & Reinforcement of the gastric pouch staple line with a suture\\
S11 &  Gastric opening & Creation of an orifice into the gastric pouch where the gastrojejunostomy will be created\\
S12 &  Exposure of the omentum & Grasping and lifting of the omentum to expose it\\
S13 &  Omental transection & Transection of the omentum to divide it into two parts\\
S14 &  Adhesiolysis & Transection of connective tissue or adhesions\\
S15 &  Treitz angle identification & Visualization of the Treitz angle to identify the proximal jejunum\\
S16 & Biliary limb measurement & Measurement of the small bowel length from Treitz angle to the future site of the gastrojejunostomy to determine the length of the biliary limb\\
S17 &  Jejunum opening & Opening of the distal jejunum where the gastrojejunostomy will be created\\
S18 & Gastrojejunal stapling & Creation of the gastrojejunostomy using a stapler \\
S19 &  Gastrojejunal defect closure & Closure of the orifice left by the stapler creating the gastrojejunostomy\\
S20 &  Mesenteric opening & Opening of the mesentery to facilitate the introduction of the stapler\\
S21 &  Jejunal transection & Transection of the jejunum proximal to the gastrojejunostomy\\
S22 &  Gastric tube placement & Movement of the gastric tube (e.g. to calibrate the size of the gastric pouch or the gastrojejunostomy)\\
S23 &  Jejunal clamping & Clamping of the jejunum distal to the gastrojejunostomy\\
S24 &  Dye injection  & Injection of dye (methylene blue) to detect any leakage of the gastrojejunostomy\\
S25 & Visual assessment & Visual inspection of the anastomosis for any leakages\\
S26 &  Gastrojejunal anastomosis reinforcement & Reinforcement of the gastrojejunostomy with an additional suture\\
S27 &  Petersen space exposure & Exposure of the Petersen space (between the alimentary limb and the transverse colon\\
S28 &  Petersen space closure & Closing of the Petersen space with suture or staples\\
S29 &  Biliary limb opening & Opening of the biliary limb where the jejunojejunostomy will be created\\
S30 & Alimentary limb measurement & Measurement of the small bowel length from the gastrojejunostomy to the future site of the jejunojejunostomy to determine the length of the alimentary limb\\
S31 &  Alimentary limb opening & Opening of the alimentary limb where the jejunojejunostomy will be created\\
S32 & Jejunojejunal stapling & Creation of the jejunojejunostomy using a stapler\\
S33 &  Jejunojejunal defect closure & Closure of the orifice left by the stapler creating the jejunojejunostomy\\
S34 &  Jejunojejunal anastomosis reinforcement & Reinforcement of the jejunojejunostomy with an additional suture\\
S35 &  Staple line reinforcement & Staple line reinforcement of the blind limb of the jejunojejunostomy with an additional suture\\
S36 &  Mesenteric defect exposure & Exposure of the mesenteric defect created by the jejunojejunostomy\\
S37 &  Mesenteric defect closure &  Closing of the mesenteric defect with suture or staples\\
S38 &  Anastomosis fixation & One or more stitches to fix the position of an anastomosis\\
S39 & Hemostasis & Any intervention to stop bleeding\\
S40 &  Irrigation/Aspiration & Irrigation and aspiration of any liquid or blood clots to be removed from the abdominal cavity\\
S41 &  Parietal closure & Closure of the abdominal wall at the trocar sites\\
S42 &  Trocar removal & Removal of all the trocars and the liver retractor\\
S43 & Suture of small bowel lesion & Rectification of a small bowel lesion using a suture\\
S44 & Drainage insertion & Insertion of drainage into the abdominal cavity to drain fluids\\
S45 & Specimen retrieval & Removal of any spare tissue (e.g. omentum, small bowel, or stomach)\\
\noalign{\smallskip}\hline
\end{longtable}

\newpage
\section{Phase and Step Hierarchy}

\begin{figure}[h!]
     \centering
     \includegraphics[angle=90, width=0.68\textwidth, clip]{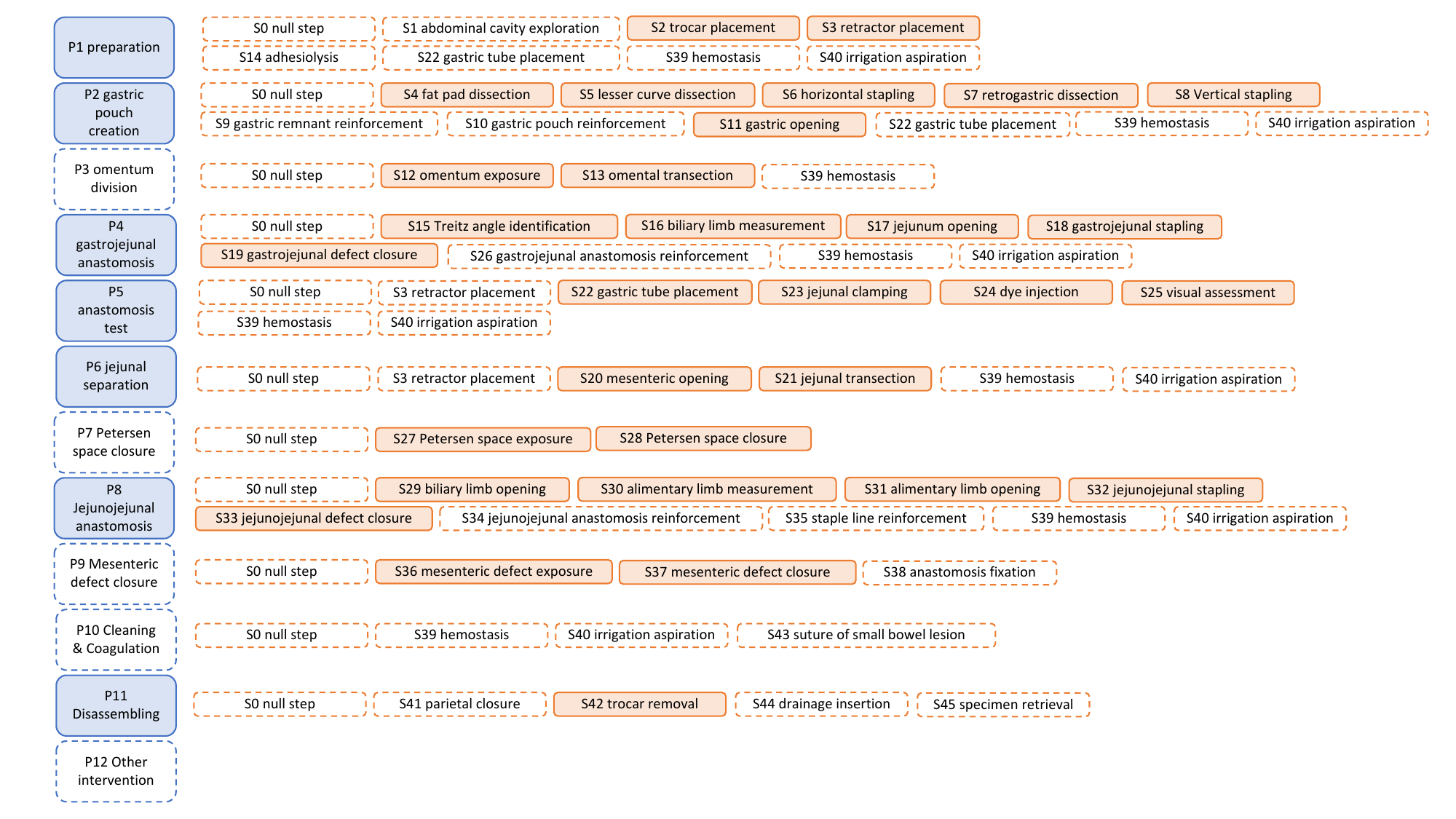}
     \caption{Hierarchical structure of phases and steps in the laparoscopic Roux-en-Y gastric bypass ontology. Facultative phases and steps have a dashed border as published in [17].}\label{ffig1:hierarchy}
\end{figure}

\newpage
\section{Dataset Characteristics}

\begin{table}[h!]
\caption{Data characteristics of the multi-center dataset presented in this work.}\label{tab1:data_characteristics}
\centering
\begin{tabular*}{0.8\textwidth}{@{\extracolsep{\fill}} cccccc}
\hline\noalign{\smallskip}
\multirow{2}{*}{Dataset} & Videos & Min. duration & Max. duration   & Mean $\pm$ std & Total\\
   & (n) & (minutes) & (minutes) & duration (minutes) & frames (n)\\
\noalign{\smallskip}\hline\noalign{\smallskip}
    StrasBypass70    \\
    Training   & 40 & 41 & 171 & $106\pm32$ & 253,913 \\
    Validation & 10 & 78 & 176 & $121\pm33$ &  72,555 \\
    Test       & 20 & 63 & 178 & $115\pm32$ & 138,326 \\
\noalign{\smallskip}\hline\noalign{\smallskip}
   BernBypass70   &&&&& \\
    Training   & 40 & 37 & 114 & $69\pm19$ & 166,431 \\
    Validation & 10 & 54 & 116 & $77\pm20$ &  46,497 \\
    Test       & 20 & 52 & 145 & $77\pm24$ &  92,979 \\
\noalign{\smallskip}\hline\noalign{\smallskip}
   MultiBypass140   &&&&& \\
    Training   & 80 & 37 & 171 & $88\pm33$ & 420,344 \\
    Validation & 20 & 54 & 176 & $99\pm35$ & 119,052 \\
    Test       & 40 & 52 & 178 & $96\pm34$ & 231,305 \\
\noalign{\smallskip}\hline
\end{tabular*}
\end{table} 

\end{document}